\documentclass[10pt,twocolumn,letterpaper]{article}

\usepackage[pagenumbers]{cvpr}              

\definecolor{cvprblue}{rgb}{0.21,0.49,0.74}
\usepackage[pagebackref,breaklinks,colorlinks,allcolors=cvprblue]{hyperref}

\usepackage{paracol}
\usepackage{makecell}
\usepackage{multirow}
\usepackage[T1]{fontenc}
\usepackage{multicol}
\usepackage{tikz}
\usepackage{xcolor}
\usepackage{float}

\usepackage[capitalize]{cleveref}
\crefname{section}{Sec.}{Secs.}
\Crefname{section}{Section}{Sections}
\Crefname{table}{Table}{Tables}
\crefname{table}{Tab.}{Tabs.}

\def\paperID{370} 
\def\confName{CVPR}
\def\confYear{2025}

\title{Harnessing Frequency Spectrum Insights for Image Copyright Protection Against Diffusion Models}

\author{
    Zhenguang Liu \textsuperscript{\rm 1 \rm2}~~ 
    Chao Shuai \textsuperscript{\rm 1 \rm2} \footnotemark[1]~~
    Shaojing Fan \textsuperscript{\rm 3} ~
    Ziping Dong \textsuperscript{\rm 1 \rm2} ~
    Jinwu Hu \textsuperscript{\rm 4}~ \\
    Zhongjie Ba \textsuperscript{\rm 1 \rm2}~ 
    Kui Ren \textsuperscript{\rm 1 \rm2} \\ 
    \textsuperscript{\scriptsize{\rm 1}}\small{The State Key Laboratory of Blockchain and Data Security, Zhejiang University,}~
    \textsuperscript{\scriptsize{\rm 2}}\small{Hangzhou High-Tech Zone (Binjiang) Institute of} \\ \small{Blockchain and Data Security,} \textsuperscript{\rm 3}\small{National University of Singapore,}~
    \textsuperscript{\rm 4}\small{South China University of Technology}~\\
    \small{\{chaoshuai, dongziping, zhongjieba, kuiren\}.zju.edu.cn},~~\{liuzhenguang2008, fhujinwu\}@gmail.com,~~dcsfs@nus.edu.sg\\
    \href{https://github.com/sccsok/CoprGuard.git}{\small https://github.com/sccsok/CoprGuard.git}
}

\begin{document}
\maketitle

\footnotetext[1]{Corresponding author}

\begin{abstract}
Diffusion models have achieved remarkable success in novel view synthesis, but their reliance on large, diverse, and often untraceable Web datasets has raised pressing concerns about image copyright protection. Current methods fall short in reliably identifying unauthorized image use, as they struggle to generalize across varied generation tasks and fail when the training dataset includes images from multiple sources with few identifiable (watermarked or poisoned) samples. In this paper, we present novel evidence that diffusion-generated images faithfully preserve the statistical properties of their training data, particularly reflected in their spectral features. Leveraging this insight, we introduce \emph{CoprGuard}, a robust frequency domain watermarking framework to safeguard against unauthorized image usage in diffusion model training and fine-tuning. CoprGuard demonstrates remarkable effectiveness against a wide range of models, from naive diffusion models to sophisticated text-to-image models, and is robust even when watermarked images comprise a mere 1\% of the training dataset. This robust and versatile approach empowers content owners to protect their intellectual property in the era of AI-driven image generation.
\end{abstract}    
\section{Introduction}
\label{sec:intro}

\par Over the past decade, we have witnessed the success of diffusion models in various fields \cite{blattmann2023stable, stablediffusion, dreambooth, xia2023diffir, saharia2022palette, chen2023single}, especially text-to-image models standing out as a prominent catalyst for stimulating creative expression. Access to high-quality training data, whether open-source or commercial, is essential to the success of diffusion models. However, this has raised a new concern over unauthorized image use in model training and fine-tuning \cite{chen2023pathway}. For example, a group of British artists have jointly filed a class-action lawsuit against Midjourney and other artificial intelligence companies \cite{guardian}. Such incidents are increasingly common and pose serious threats to both society and security. The ability to prevent image infringement during model training and fine-tuning is therefore critical in mitigating these risks.

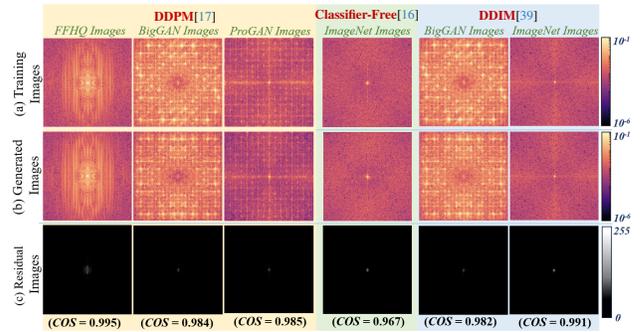
\begin{figure}[t]
    \centering
    \begin{tikzpicture}
        \node[anchor=south west, inner sep=0pt] at (0,0) {\includegraphics[width=0.47\textwidth]{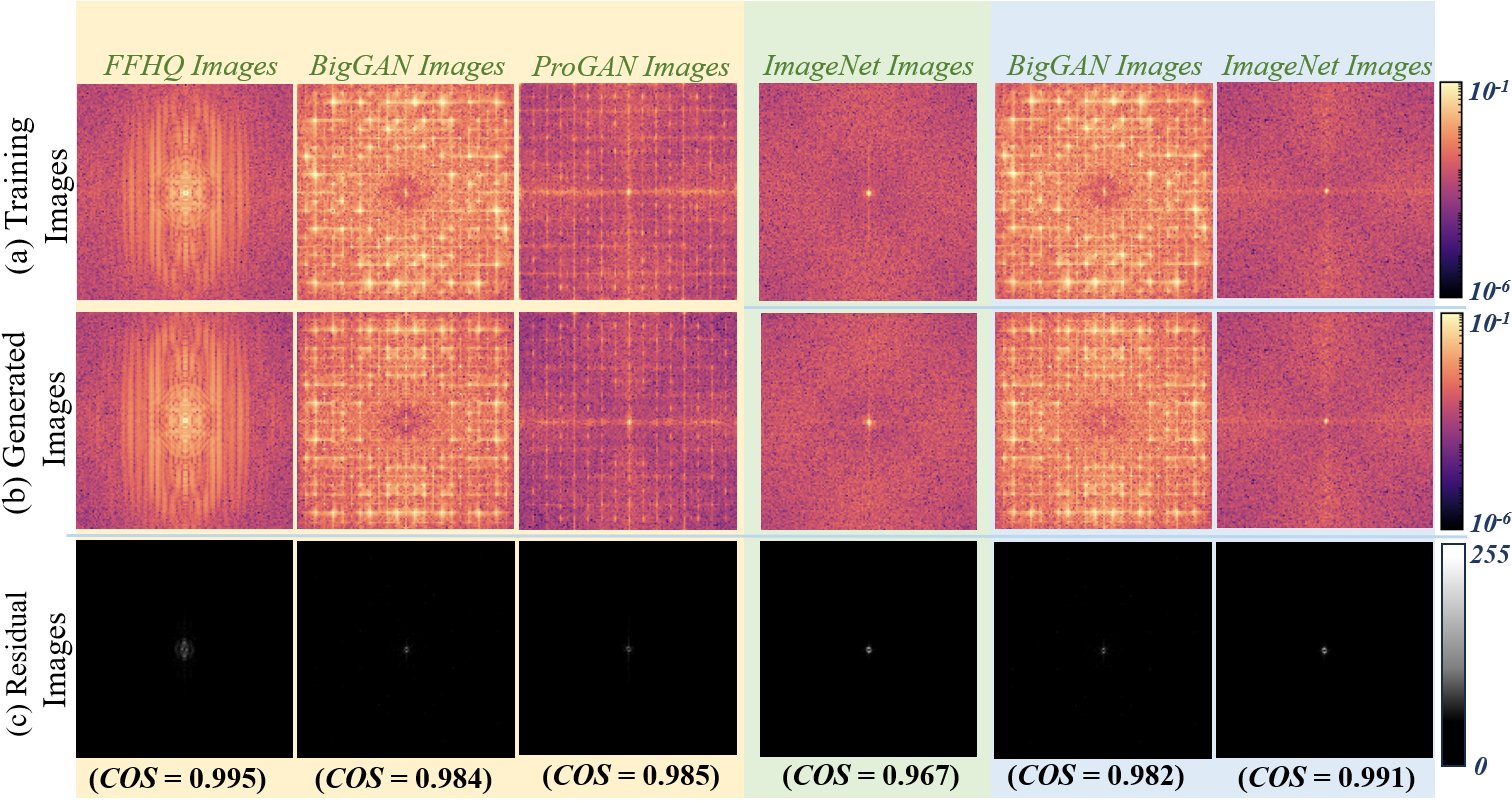}};
        \node[anchor=north east, xshift=-5.38cm, yshift=0.03cm] at (current bounding box.north east) {\textbf{{\textcolor[rgb]{0.753,0,0}{\tiny{DDPM}}}}\tiny{\cite{ddpm}}};
        \node[anchor=north east, xshift=-2.68cm, yshift=0.03cm] at (current bounding box.north east) {\textbf{{\textcolor[rgb]{0.753,0,0}{\tiny{Classifier-Free}}}}\tiny{\cite{Classifier-Free}}};
        \node[anchor=north east, xshift=-1.1cm, yshift=0.0cm] at (current bounding box.north east) {\textbf{{\textcolor[rgb]{0.753,0,0}{\tiny{DDIM}}}}\tiny{\cite{ddim}}};
    \end{tikzpicture}
    \caption{\small{Our work reveals a key property of diffusion models: generated images tend to retain the statistical properties of their training data---a finding that inspired our image copyright protection approach for diffusion models. This figure highlights the close similarity in Discrete Fourier Transform (DFT) spectra between original training images (top row) and diffusion-generated images (second row) across three diffusion models: DDPM \cite{ddpm}, DDIM \cite{ddim}, and Classifier-Free Guidance \cite{Classifier-Free}. The color scale is set to [$10^{-6}$, $10^{-1}$], and the bottom row illustrates the spectral residuals and the cosine similarities (COS)~\cite{chowdhury2010introduction}.}} 
    \label{fig1-1}
\end{figure}

Initial research has concentrated on adversarial defense methods \cite{glaze, van2023anti, liang2023adversarial}, which proactively disrupt diffusion models by adding adversarial perturbations to protect images before publication, distorting the generated outputs. While effective, these methods have key limitations: a) they lack model-agnostic flexibility, as they require tailored optimization objectives for specific models, which limits their use in black-box settings where model details are unknown, and b) adversarial perturbations are irreversible, making it impossible for authorized users to access the processed images as intended. Recent methods that use backdoors \cite{wang2023diagnosis} and watermarks \cite{zhao2023recipe, ma2023generative} offer versatility across various diffusion tasks by embedding authentication marks for infringement detection. However, they struggle to effectively support both unconditional and text-to-image models and often fail to detect infringement models when watermarked images make up a small portion of the training dataset.

To tackle these challenges, we propose a new approach to image copyright protection against diffusion models, guided by two key criteria: a) it must prevent unauthorized usage without compromising image quality or hindering model training and fine-tuning, and b) it should remain effective even if only a small fraction of the training data is unauthorized, given the mixed-source nature of diffusion model data. To achieve this, we conduct an empirical analysis of diffusion models, discovering that they retain spectral properties of their training data. This means that images generated by these models share similar frequency distributions with the original training set (see Fig. \ref{fig1-1}). Building on these insights, we introduce CoprGuard, a robust watermarking framework that embeds watermark images into the discrete wavelet transform (DWT) domain \cite{dwt} of clean images using HiNet \cite{hinet}. Additionally, we propose a watermark enhancement module within HiNet to counteract the tendency of autoencoders, such as AutoencoderKL \cite{stablediffusionv1.4} in Stable Diffusion \cite{stablediffusion}, to erase embedded watermarks.

\par To evaluate the effectiveness of CoprGuard on diffusion models (\textit{e.g.}, DDIM \cite{ddim}, Classifier-Free Guidance \cite{Classifier-Free} and Stable Diffusion \cite{stablediffusion}) and popular model training or fine-tuning methods (\textit{e.g.}, standard training and Low-Rank Adaptation (LoRA) \cite{lora}), we conduct extensive experiments on three widely used datasets: FFHQ \cite{stylegan}, ImageNet \cite{imagenet}, and Pokemon \cite{Pokemon}. Our method can achieve 100\% accuracy for the infringement model detection when the proportion of watermarked images in training dataset exceeds 5\%. CoprGuard demonstrates remarkable robustness, reliably detecting watermarked images even when they constitute as little as 1\% of the training data. Notably, it is independent of models used in the training or fine-tuning process, and has a negligible impact on the quality of training and generated images. For DDIM (trained on FFHQ) and Stable Diffusion (finetuned on Pokemon), the FID score \cite{fid} only increases by about 4 and 13 as the proportion of watermarked images in the training dataset increases from 0 to 100\%. The key contributions of this work are as follows:
\begin{itemize}
\item {To our knowledge, we are the first to uncover a fundamental property of diffusion models: generated images distribution inherits statistical characteristics, such as spectral features, from their training data. This discovery opens up new avenues for research and applications, including watermarking techniques, model analysis, and understanding the underlying mechanisms of diffusion models.}
\item {We propose a watermarking framework, CoprGuard, for image copyright protection against difussion models. It is able to detect unauthorized image usage during the training or fine-tuning for both naive and text-to-image diffusion models. A key innovation of CoprGuard is its reliance on general spectral features, making it \emph{model-agnostic}. This allows it to function effectively in black-box settings, where the model details are unknown. Additionally, CoprGuard is highly \emph{robust}, reliably extracting watermarks even when watermarked images comprise just 1\% of the training dataset. This combination of flexibility and reliability represents a significant advancement in image copyright protection against diffusion models.} 
\end{itemize}
\section{Related Work}
\label{sec:relwrk}

Diffusion models have transformed visual generation by training on vast datasets containing billions of images, but this scale has raised notable concerns about image copyright protection. In this section, we provide a brief overview of two main types of copyright protection methods: active protection using adversarial technology and passive detection.

\subsection{Adversarial Image Copyright Protection}

\par Some works \cite{glaze, van2023anti, liang2023adversarial, salman2023raising} interfere with the learning process of unauthorized generative models by proactively adding adversarial perturbations to protected images before they are published online. Shan et al. \cite{glaze} proposed the first adversarial method, GLAZE, to prevent unauthorized image usage by text-to-image models. It applies "style cloaks" (style perturbations) to artworks, misleading Stable Diffusion to unsuccessfully mimic a specific artist. Liang et al. \cite{liang2023adversarial} designed a novel method, AdvDM, which exploits adversarial samples via a Monte-Carlo estimation, preventing painting imitation by diffusion models. After that, Anti-DreamBooth \cite{van2023anti} was proposed, which adds barely perceptible perturbations to images and prevents unauthorized image usage for personalized image synthesis by DreamBooth. Salman et al. \cite{salman2023raising} obtained images that are immune to diffusion-based image editing disrupting the learning process of target diffusion models and forcing them to produce unnatural images. Shan et al. \cite{shan2023prompt} proposed Nightshade, a prompt-specific poisoning attack, which suppresses Stable Diffusion’s latest model (SDXL) from generating images with certain subjects or styles with fewer poisoned training samples. Although these methods effectively prevent images from infringement, most require pre-defined optimization objects, such as image style or subject. Besides, once modified images are released, authorized diffusion training or fine-tuning will also become impermissible.

\subsection{Passive Detection for Image Infringement}

\par Given the limitations of active protection, some studies \cite{zhao2023recipe, wang2023diagnosis, ma2023generative} opted for passive detection for image copyright protection. They  emphasize the traceability of protected images, explores detection schemes based on watermarks or backdoors, and focuses on the extraction of authentication marks from the inspected model as evidence when an image infringement incident occurs. Compared to adversarial methods, this type of method can be used to protect image copyright during the training or fine-tuning of any personalized diffusion model, and has less impact on image availability. Zhao et al. \cite{zhao2023recipe} used the pretrained watermark encoder \cite{yu2021artificial} to embed a bit string into training images, aiming to track unauthorized image usage in diffusion models. However, this method is limited to unconditional diffusion models and hardly applied to more widely used text-to-image models. Wang et al. \cite{wang2023diagnosis} proposed DIAGNOSIS, coating protected images with an image warping function \cite{nguyen2021wanet}, and determined whether text-to-image models were trained or fine-tuned with protected images by detecting if the generated image contains same features as warping image. However, their method severely degrades image quality, and when coating rate of training dataset is small, the prediction is unreliable. Ma et al. \cite{ma2023generative} proposed GenWatermark for personalized text-to-image models, but it is restricted to subject-driven synthesis. Additionally, Wu et al. \cite{wu2024cgi} leveraged reconstruction residuals of masked images to predict whether the model has been trained on a specific image, but it is limited to few-shot generation models and then examine suspected infringing images one by one.

Our approach stands apart from previous methods by its theoretical foundation. Through detailed analysis, we discovered that diffusion models tend to preserve the frequency spectrum characteristics of their training data. Building on this insight, we introduce CoprGuard, a robust watermarking framework that is highly effective for both native and text-to-image diffusion models, even when watermarked images make up only a small portion of the training data.
\section{The Spectral Signature of Diffusion Models} \label{sec3}

\par In this section, we present how we discover the fundamental property in diffusion models: the preservation of spectral information. Our experiments show that images generated by diffusion models inherit the spectral characteristics of their training data, indicating a strong link between the two. 

\par \textbf{Data preparation:} Our experiments were conducted on an image pool consisting of five subsets: 50,000 real images from FFHQ \cite{stylegan}, 50,000 images from ImageNet 1K \cite{imagenet},and 50,000 images from each of three GAN models: BigGAN \cite{biggan} pretrained on ImageNet, ProGAN \cite{progan} pretrained on LSUN, and StyleGAN2 \cite{stylegan2} pretrained on FFHQ. 

\par \textbf{Image statistics modeling:} We modeled the statistical distributions of both training data and generated images to compare their similarities, using three frequency domain transforms: the discrete Fourier transform (DFT) \cite{dft}, discrete cosine transform (DCT) \cite{dct}, and discrete wavelet transform (DWT) \cite{dwt}. All results are averaged over 20,000 images from each training and generated sub-datasets.

\par \textbf{Key Observation:} \textit{Diffusion models preserve the statistical characteristics of their training data in the generated image distribution.}

\begin{figure}[t]
    \centering
    \includegraphics[scale=0.25]{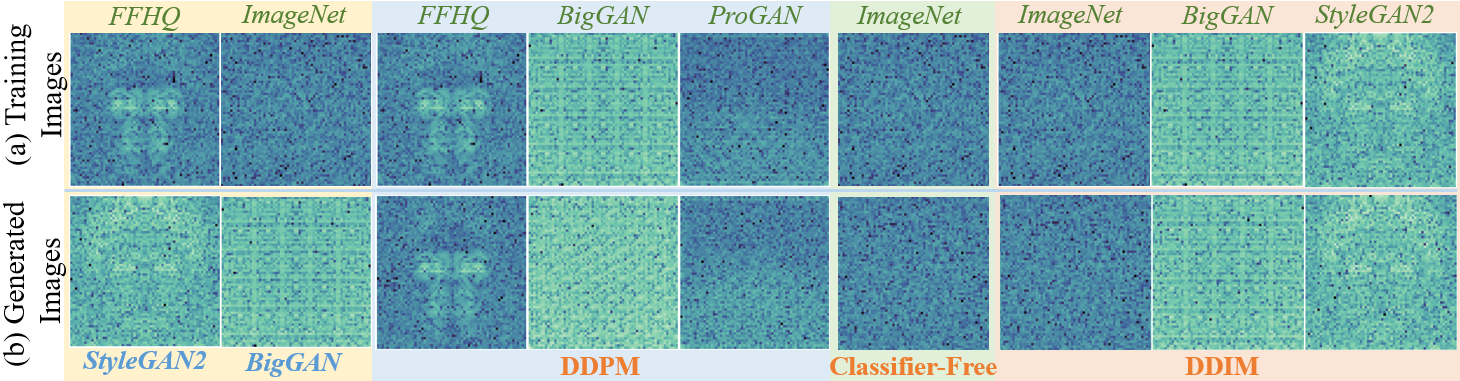}
    \caption{Mean DWT diagonal component (cD) of training images (1st row) and generated images (2nd row) for GAN and diffusion models.}
    \label{fig3-2}
\end{figure}

\begin{table}[t]
    \renewcommand\arraystretch{0.9}
    \setlength\tabcolsep{2pt}
    \caption{Cosine similarity (COS) of spectral features between diffusion-generated images and training images.}
     \small
     \centering
    \begin{tabular}{c|c|ccccccc}
    \hline
      Model & Dataset  & DFT & DCT & cA  & cD & DWT-\textit{avg} \\ \hline
      StylegGAn2 & FFHQ & 0.617 & 0.455 & 0.460 & 0.012 & 0.118 \\
      BigGAN & ImageNet & 0.564 & 0.441 & 0.443 & 0.001 & 0.119 \\ \hline
      DDIM & FFHQ & 0.988 & 0.986 & 0.987 & 0.795 & 0.937 \\
      DDPM & FFHQ & 0.995 & 0.992 & 0.991 & 0.728 & 0.891  \\
      Classifier-Freee & ImageNet & 0.967 & 0.990 & 0.991 & 0.756 & 0.824 \\
    \hline
    \end{tabular}
    \label{tab3-1}
\end{table}

\par Our experiments analyze frequency domain characteristics of generated-images by three fundamental diffusion models (DDPM \cite{ddpm}, DDIM \cite{ddim}, and Classifier-Free Guidance \cite{Classifier-Free}) for unconditional and conditional generation. We train the above three models on our five subsets, and generate 20,000 images per model with sampling step. We analyze spectrum properties of diffusion-generated images and compare them to training images. \cref{fig1-1} depicts the absolute DFT spectrum, and \cref{fig3-2} shows the DWT diagonal component (cD). 

\par Interestingly, GAN-generated images exhibit model-specific spectral features even when trained on the same dataset, essentially retaining characteristics unique to the model that generated them \cite{jeong2022frepgan, frank2020leveraging}. In contrast, diffusion-generated images consistently display spectral features closely matching those of their corresponding training datasets, even across diverse data distributions. Furthermore, \cref{tab3-1} reports the cosine similarity ($COS)$ \cite{chowdhury2010introduction} between the spectra of generated images and training images. \textbf{All spectral features for diffusion-generated images, except for DWT high-frequency components (\textit{e.g.}, cH and cD), exhibit cosine similarity scores exceeding 0.95.} The reduced similarity in high-frequency components is due to the fact that the denoising sampling of diffusion models tends to discard certain high-frequency details of the image \cite{qian2024boosting, ricker2022towards}. More spectrum visualizations (other DWT components and DCT spectra) and cosine similarities are available in Appendix \ref{C.1}. 

\begin{figure}[t]
    \centering
    \includegraphics[scale=0.22]{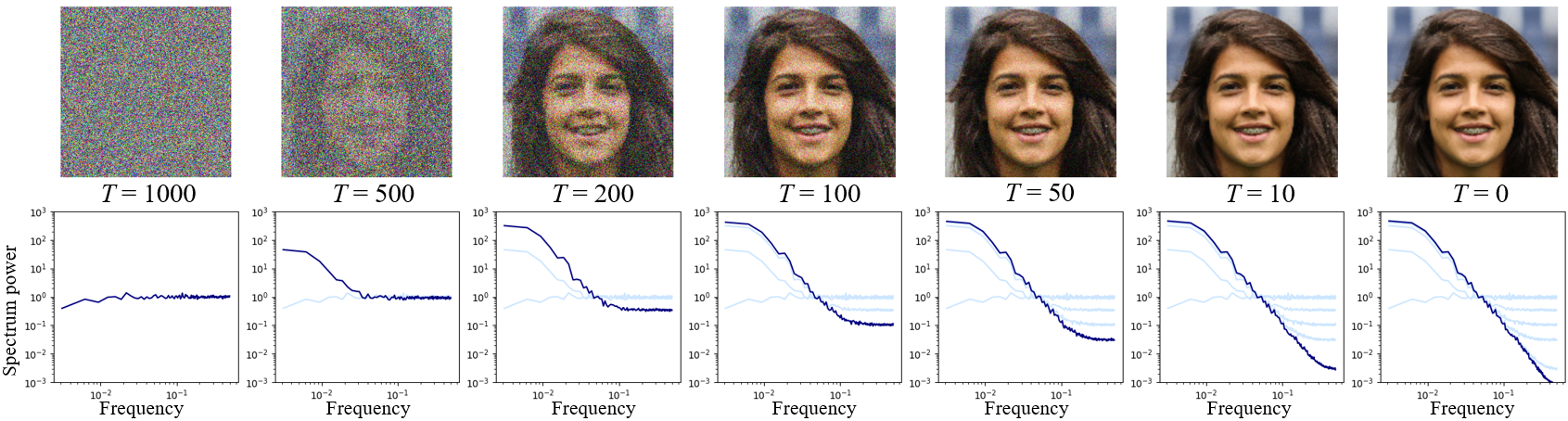}
    \caption{The RAPSD of the image with varying diffusion steps, revealing that the diffusion process progressively degrades high-frequency content and converts the image into Gaussian noise with a uniform spectral density.}
    \label{fig3-2-2}
\end{figure}

We delve deeper into our observations to explore why this occurs and seek theoretical support for our findings. Recent research has shown that diffusion models operate in the frequency domain, performing a form of approximate autoregression \cite{autoregression, rissanen2022generative, qian2024boosting}. \cref{fig3-2-2} illustrates this behavior by showing the changes in the radially averaged power spectral density (RAPSD) of images at different steps of the diffusion process. During the diffusion process, it progressively filters out high-frequency content, converting the image into Gaussian noise with uniform spectral density. Conversely, the reverse process resembles spectral autoregression.  At each step, the model predicts higher-frequency components based on lower-frequency ones, gradually reconstructing the image. This reveals a fascinating link between how diffusion models generate images and the underlying statistical properties of the training data. Essentially, the model learns to ``mimic'' the structure of the training data, even at the level of frequency components.

This observation aligns with recent findings in autoregressive models \cite{biderman2024emergent, tirumala2022memorization, tirumala2022memorization}, which demonstrate a tendency to reproduce frequently occurring elements from their training sets. This insight informs our research, leading us to hypothesize that embedding sufficient and repetitive frequency components as watermark information into the training images---without degrading image quality---will likely result in these components being preserved in the generated images. Our experiments in the following sections provide support for this hypothesis.

\section{CoprGuard: A Spectrum-Based Framework for Image Copyright Protection} \label{sec4}
\subsection{Problem Formulation}\label{sec4-1}

\par This section introduces CoprGuard, a robust image copyright protection framework designed to safeguard images from unauthorized use by diffusion models. Our primary goal is to determine whether protected images have been used in the training or fine-tuning of diffusion models.

\par We consider the protector to publish a private dataset $\mathcal{D}$, while the infringer trains a model $\mathcal{M}$ based on $\mathcal{D}$, and claims its copyright completely. The goal of the protector is to detect weather $\mathcal{D}$ is used as part of the training data during model pre-training or fine-tuning stages. Formally, the inference goal is formulated as $\mathcal{F}: \mathcal{M} \to \left\{ 0, 1 \right\}$, when the input model $\mathcal{M}$ is trained or fine-tuned based on $\mathcal{D}$, the inference result is 1; otherwise, the inference result is 0.

During the data release phase, the protector embed a watermark before releasing images to the Internet, while maintaining image quality to ensure usability. In the inspection phase, the protector has no prior knowledge of the diffusion model and only evaluates it in a black-box manner. For infringers, we consider two possible usage scenarios based on their full control over the discovered data.
\par \noindent \textit{Scenario I. } The infringer trains diffusion models from scratch with training data that originates entirely or partially from the protected dataset.

\par \noindent \textit{Scenario II. } The infringer fine-tunes the pretrained diffusion model with the protected dataset, employing full-parameter fine-tuning or parameter-efficient fine-tuning.

\begin{figure*}[t]
    \centering
    \includegraphics[scale=0.6]{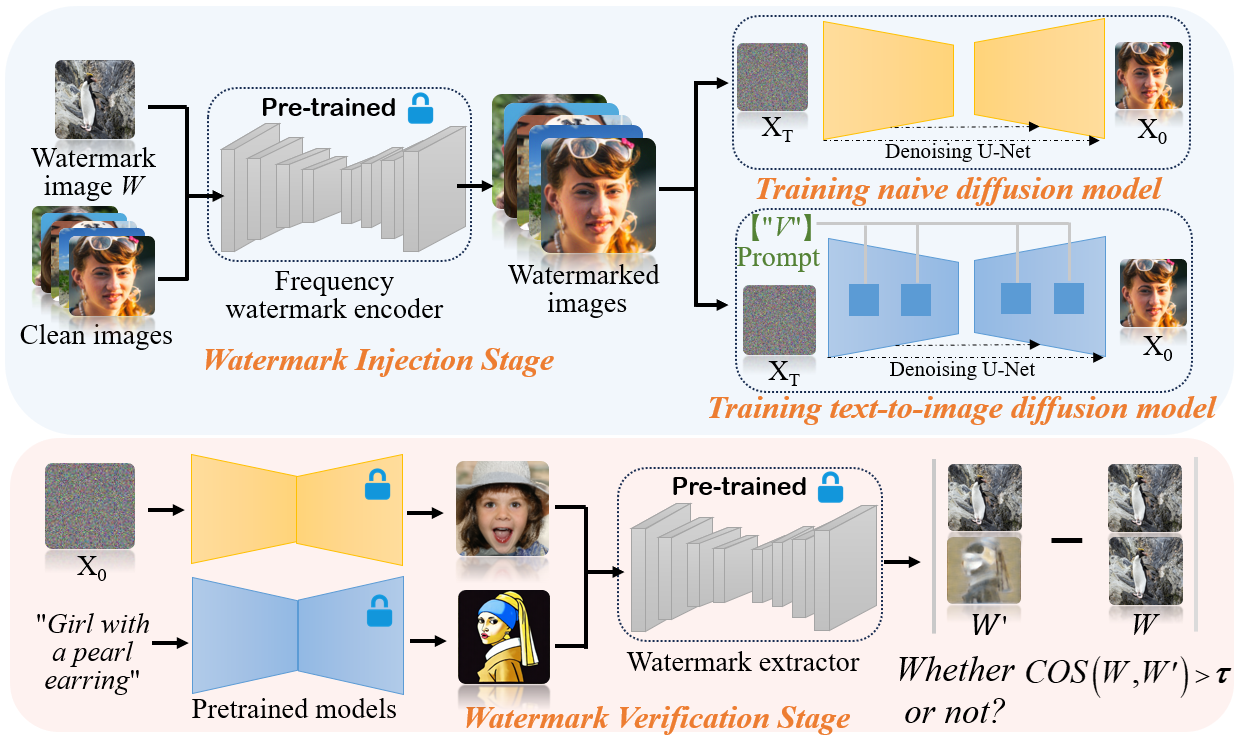}
    \caption{CoprGuard Fraemwork. The watermark image is embedded into DWT components of images using the pretrained watermark encoder, which consists of wavelet transform block and invertible neural network (INN). These images are involved into the training or finetuning of naive and text-to-image diffusion models. During inference, an image, sampled from the inspected model, is fed into the Information Enhancement Module (IEM) and pretrained watermark extractor to retrieve the watermark image $W'$. If the cosine similarity ($COS$) between the watermark $W'$ and genuine watermark $W$ exceeds the threshold $i$, the training dataset contains watermarked images.}
    \label{fig4-1}
\end{figure*}

\subsection{CoprGuard Framework Design}

\par The CoprGuard framework is illustrated in \cref{fig4-1}, and the implementation details are as follows. The protector employs a frequency domain watermark encoder to embed the watermark \textit{W} into all images before releasing the private dataset. Then if infringers train or fine-tune diffusion models (including naive and text-to-image diffusion models) with these watermarked images, the watermark \textit{W} will be embedded into models as well. When suspicious models are released, the protector generates images from these inspected models in a black-box manner and extracts the watermark $W'$ from inspected  images using the pretrained watermark decoder. Finally, the protector compares the watermark \textit{W} and $W'$ to determine whether the model uses unauthorized images and violates image copyright.

\subsubsection{Diffusion Model Watermark Injection} \label{sec4-2-1}
\par
\begin{figure}[t]
    \centering
    \includegraphics[scale=0.48]{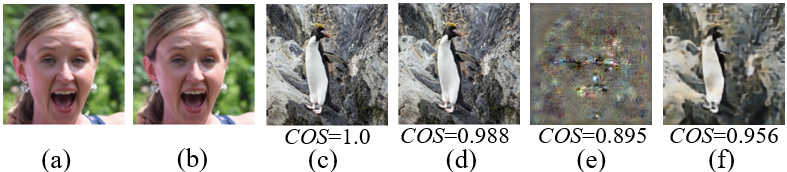}
    \caption{The customized HiNet significantly improves watermark extraction for AutoencoderKL-processed images. (a)-(c) show the clean image, watermarked image and watermark. (d) shows the extracted watermark for clean image. (e) and (f) show the extracted watermarks for AutoencoderKL-processed images by naive and customized HiNet, respectively. }
    \label{fig4-2}
\end{figure}

 Given a protected image $I$, CoprGuard first embeds a watermark image $W$ into $I$ with the frequency domain watermark method HiNet \cite{hinet}. We chose HiNet for two reasons, first, it utilizes the  wavelet transform (DWT) to split images into low and high-frequency wavelet sub-bands and embed the watermark $W$ into high-frequency components of the clear image $I$, with minimal impact on the image quality; second, we observe that the distribution learned by the autoencoder, such as the AutoencoderKL used by Stable Diffusion \cite{stablediffusion}, introduces strong frequency domain artifacts, thereby erasing watermark information. HiNet employs a learnable reversible network, allowing for  incorporating a Information Enhancement Module (IEM) before the watermark extractor to compensate for watermark degradation during encoder-decoder process.  \cref{fig4-2}(b) and (d) show the watermarked image $I_w$ and extracted watermark $W'$ for naive HiNet. \cref{fig4-2}(e) and \ref{fig4-2}(f) illustrate the extracted watermark $W'$ of AutoencoderKL-processed watermarked image $I_{rw}$, using HiNet with or without IEM, respectively.

\par The customized HiNet is described as follows: the watermark $W$ is embedded into the image $I$, producing the watermarked image $I_w$. $I_w$ is reconstructed by AutoencoderKL to simulate the image encoder-decoder process in text-to-image diffusions, yielding the reconstructed image $I'_w$. Then we input $I'_w$ into IEM, obtaining the enhanced image $I_{rw}$, which is subsequently utilized to retrieve the watermark $W'$ through the watermark extractor. When infringers train or fine-tune a diffusion model $\mathcal{M}$ with the watermark images, the protector will detect unauthorized image usage via comparing the similarity between the extracted watermark $W'$ and the original watermark $W$.

\subsubsection{Unauthorized Image Usage Detection}

\begin{table*}[t]
\caption{Effectiveness of detecting infringement diffusion models that have unauthorized image usage in model training or fine-tuning. We report CoprGuard's results with watermark injection ratio $R$ varying from 0 to 100.0\%. \textit{TN} denotes non-infringing models with $R=0$.}
\small
\centering
\renewcommand{\arraystretch}{0.85}
\setlength\tabcolsep{7pt}
\begin{tabular}{ccccccccc}
\hline
Method & Model & Dataset & Watermark Injection Ratio $R$ & $TP$ & $FP$ & $TN$ & $FN$ & Acc \\ \hline
Yu.et.al \cite{yu2021artificial} & SD + LoRA  &  Pokemon  & 25\% &  0   &  0  & 10 &  10  & 50\% \\ 
DIAGNOSIS \cite{wang2023diagnosis} &  SD + LoRA &  Pokemon  & 25\% / 100\% &  10   &  0  & 10 &  0  & 100\% \\  \hline
CoprGuard   &   DDIM   &  FFHQ  & 0  &  0   &  0  & 10 &  0  & 100\% \\ 
CoprGuard   &   Classifier-Free   &  ImageNet  &  0  &  0   &  0  & 10 &  0  & 100\% \\ 
CoprGuard   &  SD + LoRA  &  Pokemon &  0  &  0   &  0  & 25 &  0  & 100\% \\ \hline
CoprGuard   &   DDIM   &  FFHQ  & 10\% / 25\% / 50\% / 100\% &  10   &  0  & 10 &  0  & 100\% \\ 
CoprGuard   &   DDIM   &  ImageNet  &  10\% / 25\% / 50\% / 100\%  &  10   &  0  & 10 &  0  & 100\% \\ 
CoprGuard   &   Classifier-Free   &  ImageNet  &  10\% / 25\% / 50\% / 100\%  &  10   &  0  & 10 &  0  & 100\% \\ 
CoprGuard   &    SD    &  Pokemon  &  10\% / 25\% / 50\% / 100\%  &   25  &  0  & 25 &  0  & 100\%  \\ 
CoprGuard   &  SD + LoRA  &  Pokemon &  10\% / 25\% / 50\% / 100\%  &  25   &  0  & 25 &  0  & 100\% \\ \hline
\end{tabular}
\label{tab5-1}
\end{table*} 

\par In real-world scenarios, training images for diffusion models can originate from diverse sources, and often only a subset of the training data is protected (\emph{i.e.,} watermarked). This makes infringement detection considerably more difficult. In this paper, we use $R$ to represent watermark injection proportion in the training dataset, which is defined as
\begin{equation}
    {R} = \frac{{\left| {{{\mathcal{D}}_w}} \right|}}{{\left| {\mathcal{D}} \right|}},
\end{equation}
\noindent where ${\left| {\mathcal{D}}_w \right|}$ and ${\left| {\mathcal{D}} \right|}$ are the size of the watermarked subset and the whole dataset, respectively. Given a diffusion model $\mathcal{M}: \left\{ {I;p_w} \right\} \to O $ ($I$ and $O$ are the input and the output spaces), we denote ${\mathcal{O}}$ and ${\mathcal{O}}_s$ as the set of generated images and those containing the watermark $W$. Then the Watermark detection ratio $P_u$ of generated images is defined as
\begin{equation}
    {P_u} = \frac{{\left| {{{\mathcal{O}}_s}} \right|}}{{\left| {\mathcal{O}} \right|}}.
\end{equation}

\par
\begin{figure}[t]
    \centering
    \includegraphics[scale=0.47]{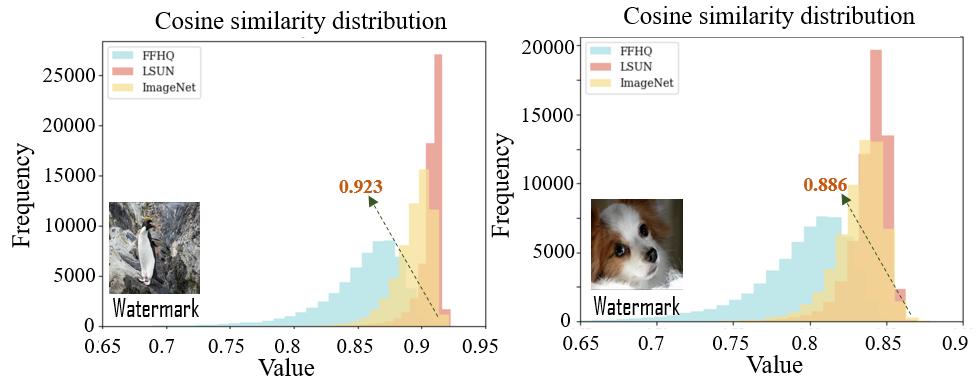}
    \caption{The cosine similarity distribution of extracted and genuine watermarks for FFHQ, LSUN, and ImageNet.}
    \label{fig4-3}
\end{figure}

 We define an image as watermarked if the cosine similarity between the extracted watermark $W'$ and genuine watermark $W$ exceeds the threshold $\tau $. Let $y$ denote the label indicating whether an image is watermarked, where $y=1$ if and only if  $COS\left( {W,W'} \right) > \tau$. The threshold $\tau$ is an empirical hyper-parameter determined through the following procedure. We select two images as watermarks, and randomly sample 50,000 clean images from FFHQ, LSUN, and ImageNet as reference images, respectively. Then we extract watermarks from reference images and compute the cosine similarity between each extracted watermark and the watermark $W$. The results are presented in \cref{fig4-3}, which demonstrates that the cosine similarities between the watermarks extracted from clean images and the watermark $W$ are below a threshold $\gamma$. Consequently, we set the threshold $\tau$ as $\tau = \gamma + 0.005$.

\par \textbf{Hypothesis Testing.} We use the statistical hypothesis testing proposed by \cite{li2023black} to predict whether $\mathcal{M}$ is trained or fine-tuned on the protected images. Let $P_w$ and $P_c$ denote the predicted probability of $y=1$ under watermarked subset $\mathcal{D}_w$ and the clean subset $\mathcal{D} - \mathcal{D}_w$, respectively. Given the null hypothesis $H_0$: $P_c + \alpha = P_w$ $\left(H_1: P_c + \alpha < P_w \right)$, where the hyper-parameter $\alpha  \in \left( {0,1} \right)$, we claim that $\mathcal{F}: \mathcal{M} \to 1$ if and only if $H_0$ is rejected. In practice, we sample $N$ images from the suspicious model $\mathcal{M}$, and with a certainty threshold $\kappa $, we can reject the null hypothesis $H_0$ at the significance level $\lambda $ if the extracted watermark rate $P_u$ satisfies that
\begin{equation}
    \sqrt {N - 1} ({P_u} - {P_c} - \kappa ) - {t_{1 - \lambda }}\sqrt {{P_u} - P_u^2}  > 0,
\end{equation}
\noindent where ${t_{1 - \lambda }}$ indicates the ${1 - \lambda }$-quantile of t-distribution with $N - 1$ degrees of freedom, and both $\kappa$ and $\lambda$ are set as the default values 0.05.

\subsection{Experiment setup}

\begin{table*}[t]
\caption{The watermark detection ratio $P_u$ of DIAGNOSIS and CoprGuard with watermark injection ratio $R$ varying from 0 to 100.0\%. }
\small
\centering
\renewcommand{\arraystretch}{0.85}
\setlength\tabcolsep{2.5pt}
\begin{tabular}{cccccccccc}
\hline
\multirow{2}{*}{Method} & \multirow{2}{*}{Model} & \multirow{2}{*}{Dataset} & \multicolumn{7}{c}{Watermark Detection Ratio $P_u$} \\ 
 &   &   & $R=0$ \color{blue}{$\downarrow$} &  $R=1\%$ \color{red}{$\uparrow$} &  $R=5\%$ \color{blue}{$\downarrow$}  &  $R=10\%$ \color{red}{$\uparrow$} & $R=20\%$ \color{red}{$\uparrow$} &  $R=50\%$ \color{red}{$\uparrow$} & $R=100\%$ \color{red}{$\uparrow$} \\   \hline
DIAGNOSIS \cite{wang2023diagnosis}  &  VQDiffusion  &  CUB-200  & 6.0\% & -- & --   &  --  & -- &  --  & 96.0\% \\ 
DIAGNOSIS \cite{wang2023diagnosis}  &  DDIM  &  FFHQ  & 3.8\% & -- & --   &  --  & -- &  --  & 100\% \\
DIAGNOSIS \cite{wang2023diagnosis}  &  SD + LoRA  &  Pokemon  & 0 & 0 &  2.0\%   &  12.0\%  & 14.0\% &  38.0\%  & 96.0\% \\ \hline
CoprGuard   &   DDIM   &  FFHQ  & 0 &1.2\% &  6.0\%   &  18.6\% & 36.2\% &  72.5\%  & 100\% \\ 
CoprGuard   &   DDIM   &  ImageNet  & 0 & 0.4\% &  5.2\%   & 14.7\%  & 30.8\% &  56.4\%  & 100\% \\ 
CoprGuard   &   Classifier-Free   &  ImageNet  & 0 & 0.6\% & 4.6\%    & 11.3\%  & 20.4\% &  47.5\% & 100\% \\ 
CoprGuard   &    SD    &  Pokemon  & 0 & 0.7\% &  7.0\%  &  16.3\%  & 30.7\% &  69.7\%  & 100\%  \\ 
CoprGuard   &   SD + LoRA  &  Pokemon & 0 & 0 & 1.7\% &  14.3\%   & 28.7\%  & 59.3\%  & 100\% \\ \hline
\end{tabular}
\label{tab5-2}
\end{table*}

\par \textbf{Implementation Details.} For \textit{scenarios I} and \textit{scenarios II} in Section \ref{sec4-1}, our experiments are conducted with three mainstream diffusion models DDIM \cite{ddim}, Classifier-Free Guidance \cite{Classifier-Free} and Stable Diffusion v1 \cite{stablediffusionv1.4}, utilizing both standard training and fine-tuning methods. FFHQ \cite{stylegan}, ImageNet 1K \cite{imagenet} and Pokemon \cite{Pokemon} are used as unconditional and text-to-image training datasets. We trained DDIM and Classifier-Free Guidance from scratch on FFHQ and ImageNet with an image size of 128. Pokemon is used to train Stable Diffusion, including full parameter fine-tuning and fine-tuning with LoRA \cite{lora}. By default, the watermarking rate $R$ is $100\%$. For unconditional generation, we sample 10,000 images per model. For Stable Diffusion, we generate 300 validation images with 60 prompts. Other hyper-parameters are discussed in Section \ref{sec4}. 

\par \textbf{Evaluation metrics. }The effectiveness of CoprGuard is measured by detection accuracy (Acc) and watermark detection rate $P_u$. Detection accuracy is defined as the ratio of correctly classified models to the total number of models evaluated. We also report True Positives ($TP$), False Positives ($FP$), False Negatives ($FN$) and True Negatives ($TN$), where the infringement model is defined as a positive sample. Additionally, we evaluate watermarked image quality using SSIM \cite{ssim} and PSNR \cite{psnr}, and calculate the FID \cite{fid} scores of the generated images.

\subsection{Evaluations}

\par \textbf{Infringement model detection performance. }In this section, we study the effectiveness of diagnosis and compare it with two methods DIAGNOSIS \cite{wang2023diagnosis} and Yu et al \cite{yu2021artificial}, which are capable of detecting unauthorized image usages in text-to-image models. 

\begin{table}[t]
\caption{The watermarked image quality of DIAGNOSIS and our method.}
\small
\centering
\renewcommand{\arraystretch}{0.85}
\setlength\tabcolsep{6pt}
\begin{tabular}{ccccc}
\hline
\multirow{2}{*}{Method} & \multicolumn{2}{c}{DIAGNOSIS \cite{wang2023diagnosis}} & \multicolumn{2}{c}{CoprGuard}  \\ \cmidrule(r){2-3} \cmidrule(r){4-5} &  PSNR &  SSIM  & PSNR  &  SSIM  \\ \hline
FFHQ    &  32.45   & 0.971 &   \textbf{}{38.02}\color{red}{$_ {5.57 \uparrow }$}    &  0.976\color{red}{$_ {0.005 \uparrow }$} \\ 
ImageNet     &   32.21  &  0.965 &    39.91\color{red}{$_ {7.7 \uparrow }$}    &  0.980\color{red}{$_ {0.15 \uparrow }$}\\ 
LSUN    &  32.84   &  0.971 &    34.93\color{red}{$_ {2.09 \uparrow }$}    &  0.963\color{blue}{$_ {0.008 \downarrow }$} \\ 
Pokemon     &   27.04  &  0.968 &    35.72\color{red}{$_ {8.68 \uparrow }$}    &  0.964\color{blue}{$_ {0.004 \downarrow }$} \\ \hline
\end{tabular}
\label{tab5-3}
\end{table}

\par We train a set of diffusion models, with or without watermarked images, using different random seeds. We then classify these models based on whether they incorporated unauthorized images during the training or fine-tuning processes. For each experiment, we train 20 pre-trained models for DDIM and Classifier-Free Guidance, and 50 pre-trained models for Stable Diffusion (SD). As shown in \cref{tab5-1}, our method successfully detects all infringement models (the watermark injection rate $R$ varying from $0$ to $100\%$), achieving an accuracy of $100\%$ for both text-to-image and naive diffusion models with $FP=0$ and $FN=0$. While the infringer collects the training or fine-tuning data from multiple sources, and the watermark injection proportion of the training dataset is 25\%, the detection accuracy for Yu et al. is only 50.0\%. Although DIAGNOSIS achieves 100\% detection accuracy at watermark injection rates of $R=25\%$ and $R=100\%$, the watermark detection ratios are $P_u=91.2\%$ for $R=100\%$ and $P_u=5.1\%$ for $R=0$ \cite{wang2023diagnosis}. In contrast, our method achieves watermark detection ratios $P_u=100\%$ for $R=100\%$ and $P_u=0$ for $R=0$.

\begin{table*}[t]
\caption{Effects of image transformations on detecting infringement diffusion models. "None" represents no image transformation.}
\small
\renewcommand{\arraystretch}{0.9}
\setlength\tabcolsep{3.6pt}
\begin{tabular}{cccccccccccc}
\hline
None & \multicolumn{2}{c}{Rotation}    & \multicolumn{2}{c}{Flip}    & \multicolumn{2}{c}{JPEG Compression}    & \multicolumn{2}{c}{Gaussian Noise}    & \multicolumn{2}{c}{Gaussian Blur}    & \multicolumn{1}{c}{Image Scaling} \\ 
- & \textit{Rot.90} & \textit{Rot.180} &\textit{Horiz} & \textit{Verti} & \textit{JPEG 70} & \textit{JPEG 50} & $\sigma  = 0.05$ & $\sigma  = 0.2$ & $k=7,\sigma = 0.5$ & $k=7,\sigma = 5$ & \textit{Crop 0.8} \\ \hline
100\% & 100\% & 100\% & 100\% & 100\% & 100\% & 100\% & 100\% & 100\% & 100\% & 100\% & 100\% \\ \hline
\end{tabular}
\label{tab:r-1}
\end{table*}

\begin{figure}[t]
    \centering
    \includegraphics[scale=0.44]{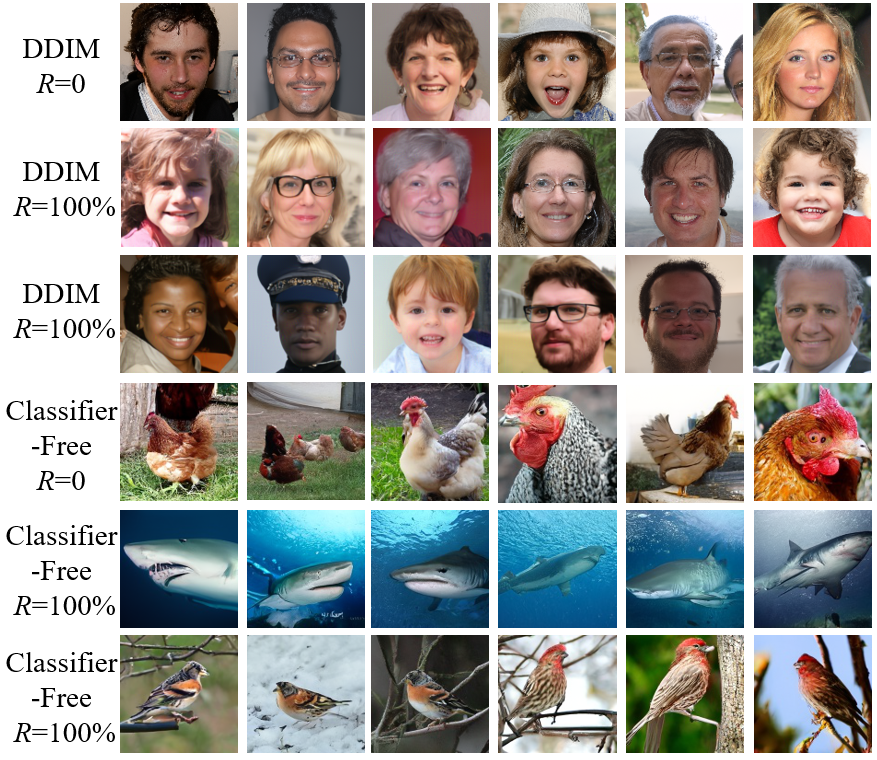}
    \caption{Generated images by DDIM and Classifier-Free Guidance. The 1st and 4th rows show generated images with watermark injection ratio $R=0$, and CoprGuard classifies all samples as clean images. The other rows show generated images with $R=100\%$, and all samples are classified as watermarked images.}
    \label{fig5-1}
\end{figure}

\begin{figure}[t]
    \centering
    \includegraphics[scale=0.27]{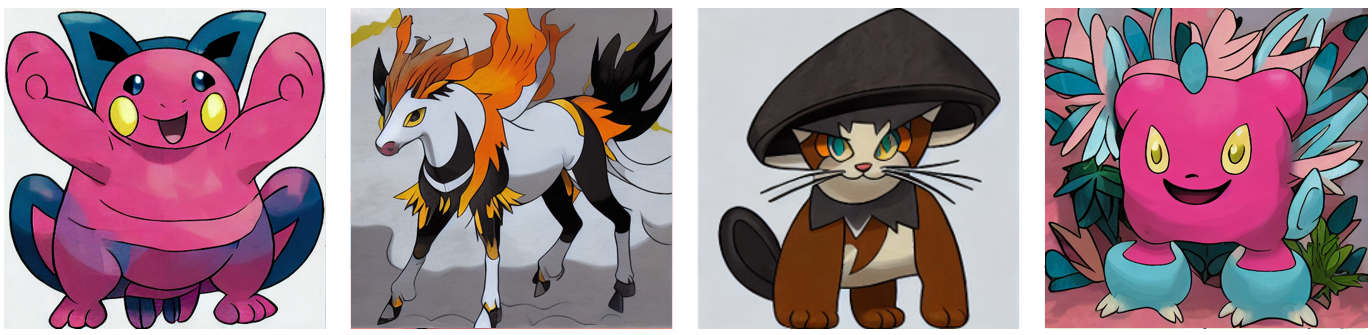}
    \caption{Generated images by Stable Diffusion with $R=100\%$, are classified as watermarked images.}
    \label{fig5-2}
\end{figure}

\par Generally, a higher watermark detection ratio $P_u$, indicates greater credibility of detection results, but $P_u$ tends to decrease as the watermark injection ratio $R$ diminishes. \cref{tab5-2} reports the watermark detection ratio $P_u$ of CoprGuard and DIAGNOSIS, with $R$ varying from 0 to $100.0\%$. When $R=0$, DIAGNOSIS still detects a small number of watermarked images from generated outputs ($P_u=6.0\%, P_u=3.8\%$), which may result in misjudgment during infringement model detection, particularly when the watermark injection ratio in the training set is low(i.e., $R<10\%$). Additionally, \cref{tab5-2} shows that, in comparison to DIAGNOSIS, our method's watermark detection ratio $P_u$ is less impacted by the injection ratio $R$. For example, in the scenario of fine-tuning with LoRA \cite{lora}, when the watermark injection ratio $R = 20\%$ and $R = 50\%$, the watermark detection ratio $P_u$ is $14\%$ and $38\%$ for DIAGNOSIS, whereas they are $28.7\%$ and $59.3\%$ for CoprGuard, indicating that our approach is more reliable as $R$ decreases.

\begin{figure}[t]
    \centering
    \includegraphics[scale=0.62]{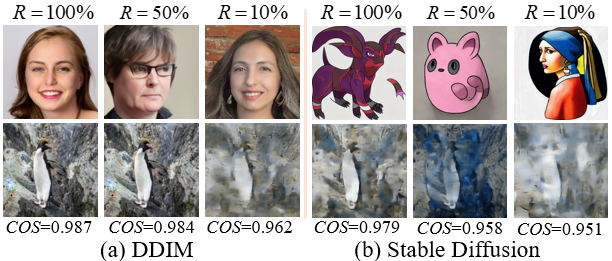}
    \caption{Extracted watermarks for images generated by DDIM and Stable Diffusion with different watermark injection ratios $R$.}
    \label{fig5-4}
\end{figure}

\begin{table}[t]
\caption{The FID scores of DDIM (trained on FFHQ) and Stable Diffusion with different watermark injection ratios $R$.}
\renewcommand{\arraystretch}{0.85}
\setlength\tabcolsep{4.5pt}
\small
\centering
\begin{tabular}{cccccc}
\hline
$R$ & 0 & 5\% & 20\% & 50\% & 100\% \\ \hline
DDIM & 14.74 & 15.12 & 16.32 & 17.35 & 18.65 \\ 
SD & 143.46  & 144.52 & 145.42 & 152.19 & 156.97 \\ 
SD + LoRA & 160.52 & 160.89 & 166.36  & 169.64 & 173.54 \\ \hline
\end{tabular}
\label{tab5-4}
\end{table}

\par \textbf{Image quality analysis. }Our goal is to protect images without affecting their legitimate use. Therefore, maintaining the quality of both training and generated images is crucial. Our investigation includes evaluating the quality of watermarked and generated images. We first analyze the impact of watermark injection on training images, with results presented in \cref{tab5-3}. Compared to DIAGNOSIS, our method exhibits a significantly smaller impact on four training datasets, achieving an average PSNR improvement of 5. We further visualize generated images from DDIM, Classifier-Free Guidance and Stable Diffusion, as shown in \cref{fig5-1} and \cref{fig5-2}. The images of the first and fourth rows in \cref{fig5-1} (sampled from models with $R=0$) are classified as clean images, while the others (sampled from models with $R=100\%$) are identified as watermarked images.

\par We further evaluated the image quality generated by three pre-trained models with different injection ratios of watermarks. The results, presented in \cref{tab5-4}, indicate that $P_u$ has minimal impact on the decline in FID scores, with a degradation of approximately 15 for Stable Diffusion under full parameter fine-tuning and fine-tuning with LoRA. 

\par \textbf{Robustness of CoprGuard. }We conducted robustness experiments on CoprGuard over common image transformations, including \textit{rotation}, \textit{flipping}, \textit{JPEG compression}, \textit{Gaussian noise}, \textit{Gaussian blur}, and \textit{image scaling}. Gaussian noise was added after normalizing images to $[0, 1]$. We trained DDIM using $50,000$ transformed images for each transformation type. \cref{tab:r-1} summarizes the watermark detection ratio $P_u$ of DDIM, trained on transformed images with a watermark injection ratio $R = 100\%$. Despite the strong impact of these transformations on the generated image quality, CoprGuard consistently achieves 100\% detection accuracy of the infringement model, demonstrating its robustness under different conditions.

\subsection{Extended analysis}

\par \textbf{Extracted watermarks of generated images. } \cref{fig5-4} (a) and (b) show the extracted watermarks for images generated by DDIM and Stable Diffusion with different watermark injection ratios $R$. Compared to DDIM, the autoencoder in Stable Diffusion degrades the quality of watermarks, even though we designed the Information Enhancement Module (IEM) to compensate for watermark information loss. Fortunately, as $R$ decreases, the generated images of both models still retain enough discernible watermarks to identify whether the model is trained based on unauthorized images.

\begin{figure}[t]
    \centering
    \includegraphics[scale=0.4]{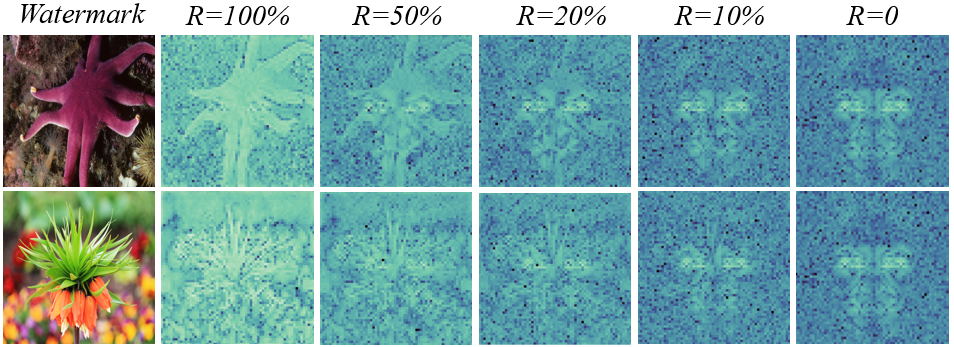}
    \caption{The mean cD components of DDIM-generated images with different watermark injection ratios $R$.}
    \label{fig5-3}
\end{figure}

\par \textbf{Frequency characteristic analysis of generated images. }In this subsection, we visualize the DWT diagonal component (cD) of DDIM-generated images with two different watermarks, as shown in \cref{fig5-3}. These figures demonstrate that watermark information is retained in the DWT spectra of generated images, supporting our hypothesis. As the watermark injection ratio $R$ decreases, the watermark information gradually degrades, but it remains detectable for CoprGuard, \textit{e.g.}, the watermark detection ratio $P_u=18.6\%$ for the watermark injection ratio $R=10\%$.

\par
\begin{figure}[h]
    \centering
    \includegraphics[scale=0.29]{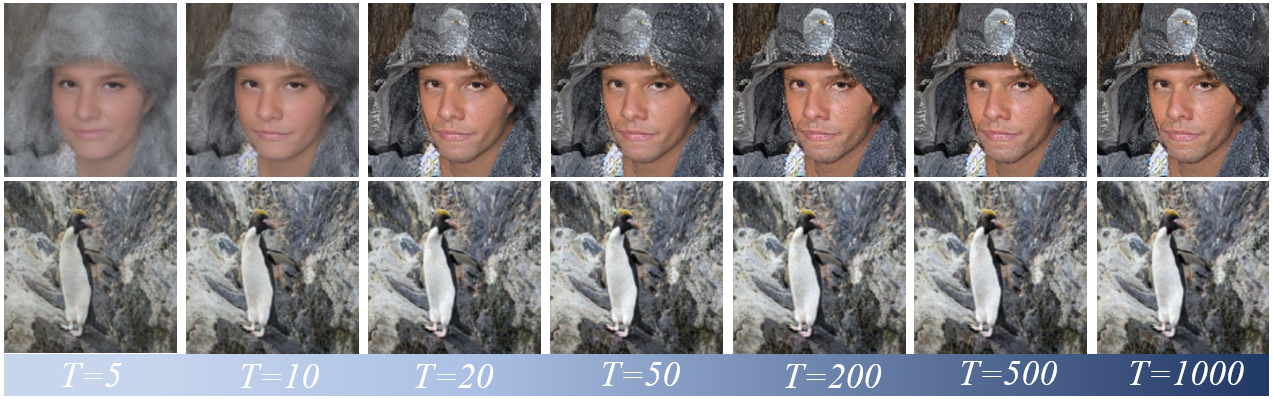}
    \caption{The extracted watermarks of DDIM-generated images with different sampling steps $T$.}
    \label{fig5-5}
\end{figure}

 \textbf{Extracted watermarks of unconditional progressive generation. }We analyze the effect of sampling step $T$ on the extracted watermarks, starting with the same initial $X_T$. The results, as shown in \cref{fig5-5}, indicate that most extracted watermarks are similar regardless of the generation trajectory. Even with $T=5$, the cosine similarity between the extracted watermark $W'$ and the original watermark $W$ is close to 1 ($COS(W,W') = 0.98)$).

\section{Conclusion and Limitation}
\par This paper uncovers a fundamental property of diffusion models: generated image distributions inherit statistical characteristics of their training data. Our primary focus is on the inheritance of spectral features, and we propose a watermarking framework, CoprGuard, to detect unauthorized image use in diffusion models. Additionally, we observe other statistical feature inheritance phenomena that can be potentially utilized, such as Photo Response Non-Uniformity (PRNU) \cite{prnu}, with detailed results provided in Appendix \ref{C.2}. Several open questions remain. First, due to the absence of training from scratch, our observations generalized to text-to-image diffusion models are not sufficient. Second, our customized HiNet still experiences watermark degradation, and we hope the advanced frequency-domain watermarking will improve this. 

{
    \small
    \bibliographystyle{ieeenat_fullname}
    \bibliography{main}
}


\end{document}